\documentclass[conference]{vldb}
\usepackage{amsfonts}
\usepackage{amsmath,amssymb,units}
\usepackage{cases}
\usepackage{mathrsfs}
\usepackage{multirow}
\usepackage{caption}
\usepackage{color}
\usepackage{flushend}
\usepackage{float}
\usepackage[ruled,vlined,linesnumbered]{algorithm2e}
\usepackage{algorithmic}
\usepackage{subfigure}
\usepackage{extarrows}
\usepackage{graphicx,graphics,txfonts}
\usepackage{array}
\usepackage{tabularx}
\usepackage{stmaryrd}
\usepackage{diagbox}
\usepackage{balance}
\usepackage{bm}
\usepackage{enumitem}
\usepackage{url}

\vldbTitle{A Sample Proceedings of the VLDB Endowment Paper in LaTeX Format}
\vldbAuthors{Jinfei Liu}
\vldbDOI{https://doi.org/10.14778/xxxxxxx.xxxxxxx}
\vldbVolume{12}
\vldbNumber{xxx}
\vldbYear{2019}

\begin{document}

\newtheorem{definition}{Definition}
\renewcommand{\algorithmicrequire}{\textbf{Requires}}
\newtheorem{theorem}{Theorem}
\newtheorem{lemma}{Lemma}
\newtheorem{assumption}{Assumption}
\newtheorem{axiom}{Axiom}
\newtheorem{example}{Example}
\newtheorem{corollary}{Corollary}
\newtheorem{property}{Property}
\newtheorem{discussion}{Discussion}
\newtheorem{remark}{Remark}

\newcommand{\partitle}[1]{\medskip \noindent \textbf{#1.}}
\newcommand{\subpartitle}[1]{\medskip \emph{#1.}}
\newcommand{\topcaption}{%
\setlength{\abovecaptionskip}{0pt}%
\setlength{\belowcaptionskip}{0pt}%
\caption}

\title{Absolute Shapley Value}

\numberofauthors{3}
\author{
\alignauthor Jinfei Liu\\
       \affaddr{Emory University and Georgia Institute of Technology}\\
       \email{jinfei.liu@emory.edu}
}

\maketitle

\begin{abstract}
Shapley value is a concept in cooperative game theory for measuring the contribution of each participant, which was named in honor of Lloyd Shapley. Shapley value has been recently applied in data marketplaces for compensation allocation based on their contribution to the models. Shapley value is the only value division scheme used for compensation allocation that meets three desirable criteria: group rationality, fairness, and additivity. In cooperative game theory, the marginal contribution of each contributor to each coalition is a nonnegative value. However, in machine learning model training, the marginal contribution of each contributor (data tuple) to each coalition (a set of data tuples) can be a negative value, i.e., the accuracy of the model trained by a dataset with an additional data tuple can be lower than the accuracy of the model trained by the dataset only.

In this paper, we investigate the problem of how to handle the negative marginal contribution when computing Shapley value. We explore three philosophies: 1) taking the original value (Original Shapley Value); 2) taking the larger of the original value and zero (Zero Shapley Value);  and 3) taking the absolute value of the original value (Absolute Shapley Value). Experiments on Iris dataset demonstrate that the definition of Absolute Shapley Value outperforms the other two definitions in terms of evaluating data importance (the contribution of each data tuple to the trained model).
\end{abstract}

\section{Background}
An acquiescent method to evaluate data importance/value to a model is leave-one-out (LOO) which compares the difference between the accuracy of the model trained by the entire dataset and the accuracy of the model trained by the entire dataset minus one data tuple \cite{DBLP:journals/pr/CawleyT03LOO}. However, LOO does not satisfy all the ideal properties that we expect for the data valuation. For example, in support vector machine (SVM), given a data tuple $p$ in a dataset, if there is an exact copy $p'$ in the dataset, removing $p$ from this dataset does not change the predictor at all since $p'$ is still there. Therefore, LOO will assign zero (or a very low) value to $p$ regardless of how important $p$ is.

Shapley value is a concept in cooperative game theory, which was named in honor of Lloyd Shapley \cite{shapley1953value}. Shapley value is the only value division scheme used for compensation allocation that meets three desirable criteria: group rationality, fairness, and additivity \cite{jia2019efficient}. Combining with its flexibility to support different utility functions, Shapley value has been extensively employed in the filed of data market \cite{agarwal2019marketplace, DBLP:conf/icml/AnconaOG19, DBLP:conf/icml/GhorbaniZ19, jia2019efficient}. One major challenge of applying Shapley value is its prohibitively high computational complexity. Evaluating the exact Shapley value involves the computation of the marginal contribution of each data tuple to every coalition, which is $\sharp P$-complete \cite{DBLP:journals/ai/FatimaWJ08}. Such exponential computation is clearly impractical for evaluating a large number of training data tuples. Even worse, for machine learning tasks, evaluating the utility function is extremely expensive as machine learning tasks need to train models. The worst case is that we need to train $O(2^n)$ models for computing the exact Shapley value for each data tuple.

A number of approximation methods have been developed to overcome the computational hardness of finding the exact Shapley value. The most representative method is Monte Carlo method \cite{DBLP:journals/cor/CastroGT09,DBLP:journals/ai/FatimaWJ08}, which is based on the random sampling of permutations. 

\section{Approximate Shapley Value}

Shapley value based compensation is a prevalently adopted approach mostly due to its theoretical properties, especially the fairness. Shapley value measures the marginal improvement of model utility contributed by $\bm{z}_i$, averaged over all possible coalitions of the data tuples. The formal Shapley value definition of $\bm{z}_i$ is shown as follows.
\begin{equation}\label{equ:shapleyValue}
  \mathcal{SV}_i=\sum_{\bm{S}\subseteq \{\bm{z}_1,...,\bm{z}_n\}\setminus \bm{z}_i}\frac{\mathcal{U}(\bm{S}\cup \{\bm{z}_i\})-\mathcal{U}(\bm{S})}{\binom{n-1}{|\bm{S}|}}
\end{equation}
where $\mathcal{U}(\cdot)$ is the utility of the model trained by a (sub)set of the data tuples, and the model utility is tested on the testing dataset.

\partitle{Monte Carlo Simulation Method}
Since the exact Shapley value computation is based on enumeration which is prohibitively expensive, we adopt a commonly used Monte Carlo simulation method \cite{DBLP:journals/cor/CastroGT09,DBLP:journals/ai/FatimaWJ08} to compute the approximate Shapley value. We first sample random permutations of the data tuples, and then scan the permutation from the first data tuple to the last data tuple and calculate the marginal contribution of every new data tuple. Repeating the same procedure over multiple Monte Carlo permutations, the final estimation of the Shapley value is simply the average of all the calculated marginal contributions. This Monte Carlo sampling gives an unbiased estimate of Shapley value. In practical applications, we generate Monte Carlo estimates until the average has empirically converged and the experiments show that the estimates converge very quickly. Therefore, Monte Carlo simulation method can control the degree of approximation, i.e., the more permutations, the better the accuracy. The detailed algorithm is shown in Algorithm \ref{Alg:MCShapley}, where $|\pi|$ is the number of permutations. The larger the $|\pi|$, the more accurate the computed Shapley value.

\begin{algorithm}[thb] \caption{Monte Carlo Shapley value computation.}\label{Alg:MCShapley}
\SetKwInOut{Input}{input}\SetKwInOut{Output}{output}

\Input{$\bm{Z}_{train}=(\bm{X}_{train},\bm{y}_{train})$ and $\bm{Z}_{test}=(\bm{X}_{test},\bm{y}_{test})$.}
\Output{Shapley value $\mathcal{SV}_i$ for each data $\bm{z}_i$ in $\bm{Z}_{train}$.}

initialize $\mathcal{SV}_i=0$\;

    \For{k=1 to $|\pi|$}{
        we have a training dataset ordered in $\pi^k$, $\bm{Z}_{train}^k=\{\bm{z}_{\pi^k_1},\bm{z}_{\pi^k_2},...,\bm{z}_{\pi^k_n}\}$\;
        \For{i=1 to n}{
            $\mathcal{SV}(\bm{z}_{\pi^k_i})=\mathcal{U}(\{\bm{z}_{\pi^k_1},...,\bm{z}_{\pi^k_i}\})-\mathcal{U}(\{\bm{z}_{\pi^k_1},...,\bm{z}_{\pi^k_{i-1}}\})$\;
            $\mathcal{SV}_{\pi^k_i}=\mathcal{SV}_{\pi^k_i}+\mathcal{SV}(\bm{z}_{\pi^k_i})$\;
        }
    }
\end{algorithm}

\section{Shapley Value Definitions}

\begin{figure}[htb]
 \centering
 \includegraphics[width=0.5\textwidth]{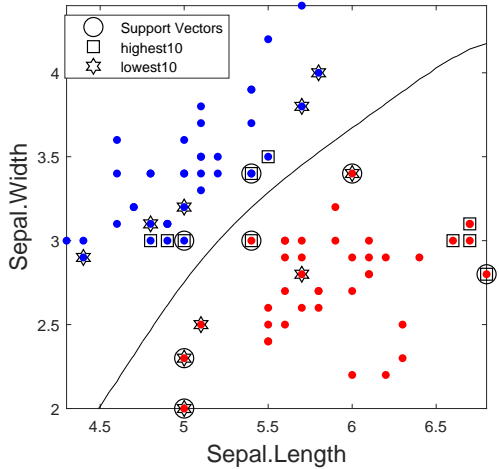}
 \vspace{-1em}
 \caption{Original Shapley Value}
 \label{fig:SVwith}
\end{figure}

The existing work \cite{DBLP:conf/icml/AnconaOG19, DBLP:conf/icml/GhorbaniZ19} takes the original value when computing the marginal contribution of each data tuple to each coalition. The formal definition named as \emph{Original Shapley Value} is shown in Equation (\ref{equ:shapleyValue}). A toy experiment result is shown in Figure \ref{fig:SVwith}. For the ease of visualization, we take the first two features (sepal length and sepal width) and the first two species (Iris Setosa and Iris Versicolour) from the classic Iris dataset \cite{Dua:2019}. We randomly choose 20 data tuples as the testing dataset and the remaining as the training dataset. Iris Setosa (Versicolour) is shown in red (blue) color. The model utility $\mathcal{U}(\cdot)$ is measured by SVM. The data tuples in support vectors are shown in circle, and the top 10 data tuples with the highest (lowest) Shapley value are shown in square (pentagram) denoted by highest10 (lowest10).

\begin{figure}[htb]
 \centering
 \includegraphics[width=0.5\textwidth]{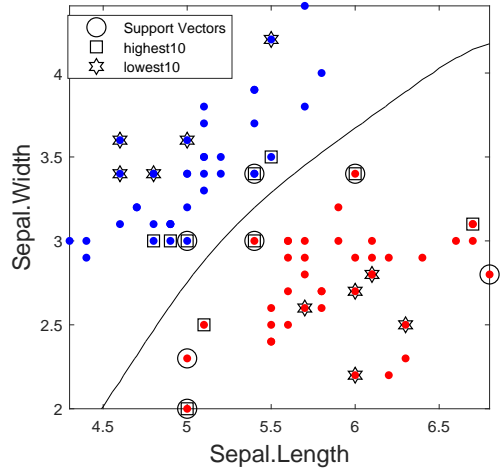}
 \vspace{-1em}
 \caption{Zero Shapley Value.}
 \label{fig:SV0}
\end{figure}

Alternatively, if the marginal contribution of a data tuple to a coalition is a negative value, we may take zero rather than the original value. The formal definition named as \emph{Zero Shapley Value} is shown in Equation (\ref{equ:shapleyValue0}) and the toy experiment result is shown in Figure \ref{fig:SV0}.

\begin{equation}\label{equ:shapleyValue0}
  \mathcal{SV}_i=\sum_{\bm{S}\subseteq \{\bm{z}_1,...,\bm{z}_n\}\setminus \bm{z}_i}\frac{\max\{0,\mathcal{U}(\bm{S}\cup \{\bm{z}_i\})-\mathcal{U}(\bm{S})\}}{\binom{n-1}{|\bm{S}|}}
\end{equation}

\begin{figure}[htb]
 \centering
 \includegraphics[width=0.5\textwidth]{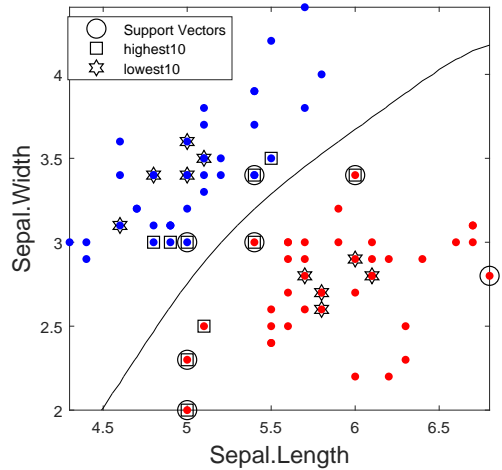}
 \vspace{-1em}
 \caption{Absolute Shapley Value.}
 \label{fig:SVabs}
\end{figure}

\textcolor{red}{We observe that whether the effect of adding a new data tuple is positive or negative, as long as the effect is large enough, the newly added data tuple is significant to the trained model.} Therefore, we propose a new definition named as \emph{Absolute Shapley Value}. In absolute Shapley value, if the marginal contribution of a data tuple to a coalition is a negative value, we take its absolute value. The formal definition is shown in Equation \ref{equ:shapleyValueabs} and the toy experiment result is shown in Figure \ref{fig:SVabs}.

\begin{equation}\label{equ:shapleyValueabs}
  \mathcal{SV}_i=\sum_{\bm{S}\subseteq \{\bm{z}_1,...,\bm{z}_n\}\setminus \bm{z}_i}\frac{|\mathcal{U}(\bm{S}\cup \{\bm{z}_i\})-\mathcal{U}(\bm{S})|}{\binom{n-1}{|\bm{S}|}}
\end{equation}

\partitle{Observation}
Shapley value is employed to evaluate data importance to a model. Therefore, the more important the data tuple, the higher the Shapley value. In SVM classifier model, generally speaking, the data tuples in the support vectors should have higher Shapley value. In original Shapley value of Figure \ref{fig:SVwith}, highest10 contains only 4 data tuples in the support vectors. Even worse, lowest10 also contains 3 data tuples in the support vectors. Differently, in zero Shapley value of Figure \ref{fig:SV0}, highest 10 contains 5 data tuples in the support vectors. Absolute Shapley value has the best performance in which highest10 contains 6 data tuples in the support vectors. Furthermore, lowest 10 in absolute Shapley value is more compact and in the middle than lowest10 in zero Shapley value, i.e., the data tuples in lowest10 of absolute Shapley value are more unimportant than the data tuples in lowest10 of zero Shapley value.

\section{Experiment}
We ran experiments on a machine with an Intel Core i7-8700K and two NVIDIA GeForce GTX 1080 Ti running Ubuntu with 64GB memory. We compute Shapley value of each data tuple based on the following definitions in Python 3.6.

\begin{itemize}
\item \textbf{ORI}: Original Shapley value definition in Equation (\ref{equ:shapleyValue}).

\item \textbf{ZERO}: Zero Shapley value definition in Equation (\ref{equ:shapleyValue0}).

\item \textbf{ABS}: Absolute Shapley value definition in Equation (\ref{equ:shapleyValueabs}).

\end{itemize}

\subsection{Performance on Iris dataset}
\label{sec:iris}
The Iris flower dataset \cite{Dua:2019} is a multivariate dataset introduced by the British statistician and biologist Ronald Fisher, which consists of 50 samples from each of three species of Iris (Iris setosa, Iris virginica, and Iris versicolor). Four features are sepal length, aepal width, petal length, and petal width, in centimeters. 

We employ two classic machine learning models, Logistic Regression (LR) and Support Vector Machine (SVM), to evaluate the effectiveness of different Shapley value definitions. We first compute Shapley value of each data tuple using the Monte Carlo method in the training dataset. We then train two predictive models from scratch based on the top $K$ training data tuples with the highest Shapley value and the top $K$ training data tuples with the lowest Shapley value, respectively. 

The model accuracy is reported in Table \ref{table:iris}. Surprisingly, for both LR and SVM, the accuracy of the model trained by the top K training data tuples with the highest ORI equals to the accuracy of the model trained by the top K training data tuples with the lowest ORI. However, this phenomenon exactly validates our guess that ``whether the effect of adding a new data tuple is positive or negative, as long as the effect is large enough, the newly added data tuple is significant to the trained model''. Recall Figure \ref{fig:SVwith}, not only the data tuple in highest10 can be contained in support vectors, but also the data tuple in lowest10. Furthermore, for both LR and SVM, the model trained by the top K training data tuples with the lowest ABS has the lowest accuracy, which verifies that the data tuples with the lowest ABS are truly unimportant. Recall Figure \ref{fig:SVabs}, lowest10 data tuples lie in the middle of the dataset, and they are unimportant to the model accuracy. Therefore, ABS outperforms ORI and ZERO in terms of evaluating data importance.

\begin{table}[htb]
\centering
\caption{Model accuracy ($K=35$) on Iris dataset. }
\begin{tabular}{c|c|c|c|c}
& LR  & LR  & SVM  & SVM  \\ 
&  (highestK) & (lowestK) & (highestK) & (lowestK) \\ 
\hline 
ORI  & 100.00\% & 100.00\% & 93.33\% & 93.33\% \\ 
ZERO & 100.00\% &  63.33\% & 96.66\% & 90.00\%  \\ 
ABS  & 100.00\% &  60.00\% & 96.66\% & 90.00\%\\ 
\end{tabular}
\label{table:iris}
\end{table}

\section{Conclusion and Future Work}
In this paper, for the first time, we define absolute Shapley value for evaluating data importance in training machine learning models. The experimental results of LR and SVM on Iris dataset show that absolute Shapley value definition outperforms original Shapley value and zero Shapley value in terms of evaluating data importance. For future work, we would like to explore the effectiveness of different Shapley value definitions on more machine learning models and more representative datasets.
\\
\\

\bibliographystyle{plain}
\bibliography{CDP}
\end{document}